\RequirePackage{fix-cm}
\documentclass{aipher_paper}
\newif\ifzh \zhfalse
\usepackage{fontspec}

\usepackage{longtable}
\usepackage{amsmath,amssymb,array,tabularx,enumitem,fancyvrb,needspace}
\usepackage[expansion=false]{microtype}
\hypersetup{bookmarksdepth=2,bookmarksopen=false}
\setlist{nosep,leftmargin=1.4em}
\ifzh
  
\else
  
\fi
\newcommand{\paperTitle}[1]{\title{#1}\hypersetup{pdftitle={#1}}}
\newcommand{\paperAuthorNames}{}
\newcommand{\paperAuthor}[2]{%
  \author[#2]{#1}%
  \ifdefempty{\paperAuthorNames}{}{\appto\paperAuthorNames{, }}%
  \appto\paperAuthorNames{#1}%
  \hypersetup{pdfauthor={\paperAuthorNames}}}
\renewcommand{\affiliationfont}{\fontsize{9}{11}\selectfont}
\renewcommand{\affiliation}[2][]{%
  \ifdefempty{\affiliationlist}{}{\appto\affiliationlist{\\}}%
  \appto\affiliationlist{\affiliationformat[#1]{#2}}}
\newcolumntype{Y}{>{\raggedright\arraybackslash}X}
\newcommand{\code}[1]{\hspace{0pt}{\small\nolinkurl{#1}}\hspace{0pt}}
\makeatletter\let\tableinput\@@input\makeatother

\paperTitle{\texorpdfstring{UMR: A Lean-Style Language for Representation\\and Structural Verification of Organic Reaction Mechanisms}{UMR: A Lean-Style Language for Representation and Structural Verification of Organic Reaction Mechanisms}}
\paperAuthor{Sirui Shao}{1,2}
\paperAuthor{Haichuan Tan}{1,3,4}
\paperAuthor{Tao Huang}{1,5}
\paperAuthor{Bowen Gao}{1,3,4}
\affiliation[1]{AIPher Technology}
\affiliation[2]{Department of Chemistry, Zhili College, Tsinghua University}
\affiliation[3]{Institute for AI Industry Research (AIR), Tsinghua University}
\affiliation[4]{Department of Computer Science and Technology, Tsinghua University}
\affiliation[5]{School of Pharmaceutical Sciences, Peking University}
\abstract{Explaining an organic reaction requires an account of how electron movements connect the reactants to the product under the stated conditions. A large language model can give a coherent account of how a known product forms even when the proposed electron movements fail to connect the intermediates it lists. Checking the product alone can miss such a spurious mechanism. We introduce Unified Mechanism Representation (UMR), an executable language in which polar, radical, pericyclic and organometallic processes share a state and event representation. The model proposes electron movements on the current Lewis structure, and the executor uses them to reconstruct the next state. The executor can independently verify the program by replaying it from the initial state and checking electron and material balance at each step. If a check fails, the executor identifies where and why, so the model can revise the path from the preceding intermediate. We find that UMR substantially improves GPT-6.2 Astra's ability to infer basic reaction mechanisms. We compared two ways of solving the 319 known-product problems in FukuyamaBench, giving the model the same substrates, products and conditions in both settings. When the model submitted intermediates directly, it matched every reference structural checkpoint in order on 62 problems (19.4\%). With UMR and its executor, it did so on 83 (26.0\%). Of the net gain of 21 problems, 14 came from the introductory set, five from the graduate-level set and two from the research-laboratory set.}

\begin{document}
\maketitle
\section{Introduction}
Even when both the starting materials and the product of a reaction are known, what happens between them is a chemical question. In Figure~\ref{fig:events}a, nitrogen in ammonia donates a lone pair to the carbon of methyl bromide as the original carbon--bromine bonding pair moves to bromine. Drawing both arrows in the same step explains how the carbon--nitrogen bond can form and bromide can leave. In a multistep reaction, the intermediate formed at each step supplies the starting point for the next. Chemists use these connections to propose testable mechanisms. Kinetic experiments, isotope labelling or stereochemical observations can then distinguish routes to the same product.\cite{grossman,mechanism}

For a language model, a known product supplies a goal and can also conceal a hallucination. A model may recognise a familiar transformation, name plausible steps and work backwards from the product to give a fluent explanation. Even when each intermediate looks reasonable on its own, the electron movements the model describes may fail to produce the next intermediate from the species present at the preceding step. We call such an account, whose narrative is coherent but whose structures do not connect as claimed, a \emph{spurious mechanism}.

Existing representations address different parts of this task. SMILES records the molecular graph of each intermediate. Two successive graphs alone leave the electron movements unspecified.\cite{smiles} FlowER uses bond--electron matrices to predict electron redistribution under constraints that conserve atoms and electrons.\cite{flower} MechSMILES appends curved-arrow instructions to SMILES strings. Its published workflow is limited to closed-shell polar mechanisms and does not give the model real-time feedback with which to revise a failed step.\cite{mechsmiles} None of these published methods provides a common executor for polar, radical, pericyclic and organometallic events that can check the model's proposed actions step by step, identify a failed event and let the model revise it from the same starting state.

Mathematical reasoning illustrates the same difficulty. A model may reach the right conclusion in a proof that relies on a premise it has yet to establish. Such a spurious proof can read like a conventional argument. Checking the last line misses the inference on which the argument breaks down.\cite{proofbench} Lean offers a way to address this problem. Once the statement and its assumptions are formalised, a model can inspect the current goal, try a proof step and use the remaining goals or an error message to revise its argument. A completed proof supplies a term that an independent kernel checks against the type of the proposition under the stated assumptions. The model that constructs the proof can thus rely on an independent check of its reasoning.\cite{lean,leandojo}

We introduce Unified Mechanism Representation (UMR) to bring stepwise construction and independent verification to organic mechanisms. UMR assigns persistent identities to atoms and records the distribution of electrons in the current state as a Lewis structure. The model describes a change as an elementary event. It specifies which electrons move, names their sources and destinations, and groups moves that must occur together. The executor reads the species, bonds and electrons present in the preceding state, checks that the moves can be made together, and reconstructs the resulting structure. If an event fails, the model can inspect the diagnostic and revise its proposal at the intermediate it has already reached. The executor can also draw a mechanism diagram automatically from the sequence of accepted actions for a chemist to review.

Polar, radical, pericyclic and organometallic processes share the same state and event representation. Resonance contributors are recorded as different Lewis views of one state. Replay establishes the structural consequences of the proposed events under the stated rules. When multiple pathways are executable, deciding which prevails under experimental conditions still requires evidence about selectivity.

The same mechanism can be read at several levels. A chemist can start with a named reaction, expand it into basic transformations and trace an individual bond back to the electron movements that form it. A model can revise one elementary event and then continue along the pathway.

We test this approach on the 319 known-product problems in FukuyamaBench. Given the same starting materials, conditions and final product, GPT-6.2 Astra either submits the intermediates directly or proposes events in UMR, inspects the resulting states or diagnostics, and selects a path. Both settings use four demonstrations matched for chemical content and one independent attempt per problem. The reference intermediates are hidden during inference. The same official scorer awards credit when a submission matches every reference structural checkpoint in order. Complete matches rise from 62/319 (19.4\%) to 83/319 (26.0\%). The introductory set accounts for 14 of the net gain of 21 (23/78 to 37/78). The graduate-level and research-laboratory sets contribute net gains of five and two, respectively. Figure~\ref{fig:score-comparison} compares the results by partition with the published benchmark.\cite{fukuyama}

We present the UMR language and its executor for writing organic mechanisms as programs that can be expanded at several levels of detail and verified step by step by a third party. We evaluate how these programs support a language model during mechanism inference. Examples drawn from two named-reaction textbooks illustrate this connection across levels of description. The paired evaluation shows what the language and its executor offer during inference: allowing the model to construct and revise an executable pathway improves complete reference-pathway accuracy on this known-product task, with the largest gain on introductory problems. The UMR language, structural checker, executor, and named-reaction programs are available at \url{https://github.com/SrCO3/UMR-Language}.

\section{Foundations of the UMR Language}
\subsection{Lewis structures, closed shells and element-specific rules}
\begin{figure}[!htbp]\centering
\includegraphics[width=\linewidth]{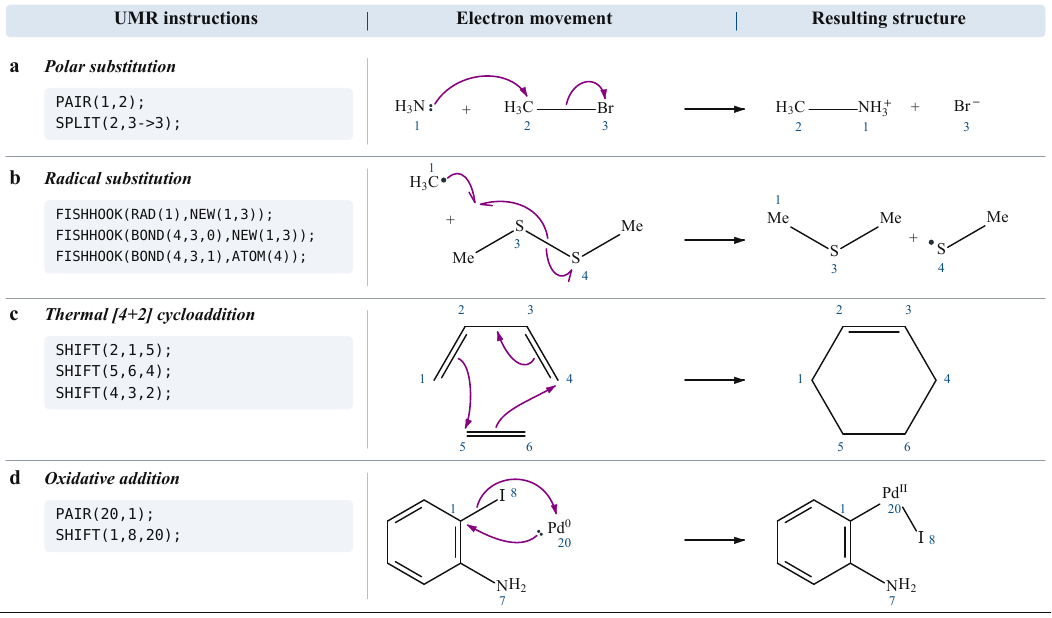}
\caption{Four elementary events, each read from UMR instructions to electron movements and the resulting state. (a) Ammonia displaces bromide. (b) Radical substitution. (c) Three electron pairs move in one thermal [4+2] event. (d) Oxidative addition of an aryl iodide to Pd(0). Numbers identify the same atoms before and after each event. A curved arrow with a half head shows one electron, and a full head shows an electron pair. Me denotes methyl.}\label{fig:events}
\end{figure}

A single covalent bond in a Lewis structure represents a shared electron pair. Atoms may also carry lone pairs or unpaired electrons. For an ordinary main-group covalent site, let $v_i$ denote the valence-electron count of the neutral atom, $n_i$ its nonbonding electrons and $B_i$ the sum of bond orders including attached hydrogens. Formal charge and shell occupation are then
\[
       q_i=v_i-n_i-B_i,\qquad N_i=n_i+2B_i .
\]
In a closed-shell structure, electrons occupy orbitals in pairs. The octet rule provides a local check for many closed-shell main-group sites. The carbon in Figure~\ref{fig:events}a exchanges one neighbour and retains eight shell electrons. Hydrogen has a full shell with two electrons. A carbon with three single bonds and zero, one or two nonbonding electrons is a cation, a radical or an anion, respectively.\cite{lewis,langmuir}

UMR applies these rules according to the element and its local electronic configuration. A neutral three-coordinate boron can have six shell electrons, and some closed-shell structures of phosphorus, sulfur and the heavier halogens exceed an octet. At metal centres, UMR separately records and checks the oxidation state, ligands and electron count. In Appendix A, Figure~\ref{fig:umr-element-scope} shows element coverage, and Table~\ref{tab:element-registry} lists the validation route and oxidation states for each registered element.

\subsection{States, views and event execution}\label{sec:views}\label{app:res}
The tail of a curved arrow marks the source of electrons and its head marks their destination. A full-headed arrow moves a pair and a fishhook moves one electron. An \emph{elementary reaction} in a proposed mechanism proceeds in one step without an intervening intermediate. It may require several simultaneous arrows, and a laboratory operation may encompass several such steps. UMR groups the actions that read one preceding state into a single event. When a source treats two processes together, an event can record that choice of resolution.\cite{iupac,elementary}

Write $S$ for a chemical state with its atom identities and material inventory. A Lewis view $V$ of that state is one assignment of bond orders, nonbonding electrons and formal charges to specific bonds and atoms. Executing a \code{REACT} event $e$ yields a successor state. With $\Gamma$ denoting the current stage, local rules and coordination declarations, a successful transition is written
\[
            \Gamma\vdash(S,V)\xrightarrow{\,e\,}(S',V').
\]
Resonance changes the Lewis view of one state. Appendix~\ref{app:resonance} follows the electrons through both contributors of an enolate.

\FloatBarrier\Needspace{7\baselineskip}
\section{From elementary events to named reactions}
\subsection{Units of composition}
At the lowest level an action names the source and destination of a quantity of electrons. An \emph{elementary event} groups actions on one pre-state with the preconditions they must jointly meet. This lets the executor detect an omission or a conflict within the group. A \emph{basic transformation} separates chemical roles from the atom indices of any one example. In Figure~\ref{fig:events}a, N1, C2 and Br3 occupy the roles of nucleophile, electrophilic centre and leaving group. Those roles and actions can be rebound to another substrate satisfying the preconditions. Here the basic transformation requires one event. Others require several. Unimolecular nucleophilic substitution ($\mathrm{S_N1}$) and unimolecular elimination (E1), for example, may both begin with ionisation. A nucleophile then captures the intermediate in substitution, whereas loss of a proton from an adjacent carbon gives elimination. A \emph{named reaction} composes basic transformations and passes intermediates, materials and catalyst states from one to the next. A reader can move from the name used in a synthetic scheme down to the source of the electrons in a particular bond.

\subsection{Unfolding a catalytic pathway}
\begin{figure}[!htbp]\centering
\includegraphics[width=\linewidth]{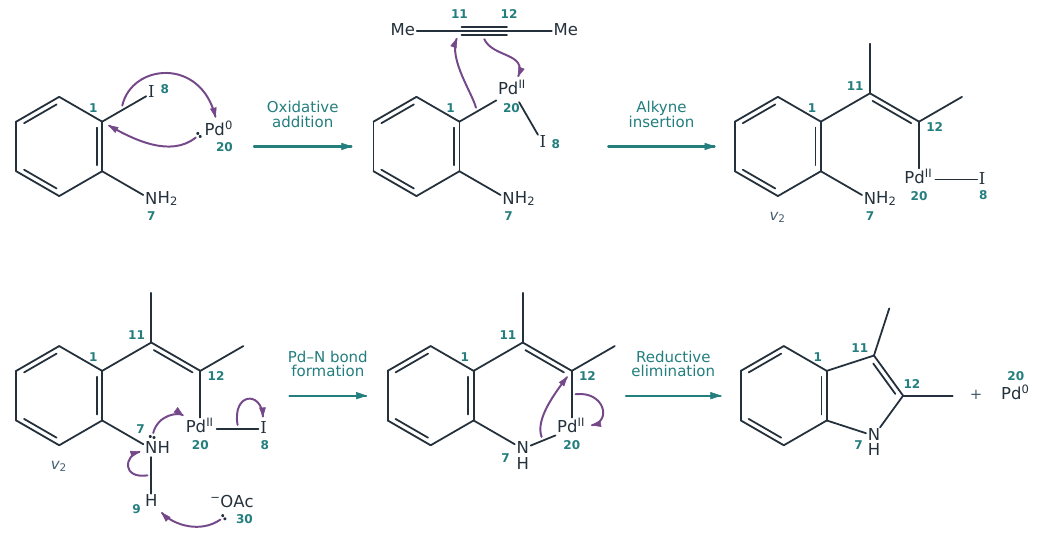}
\caption{Four events in the Larock indole synthesis. Read each row from left to right. The vinylpalladium intermediate $v_2$ is repeated at the line break. The lower row shows the reacting N--H bond and acetate. C12--N7 bond formation regenerates Pd(0). Me denotes methyl and OAc acetate. Two PPh$_3$ ligands remain coordinated to Pd throughout and are omitted for clarity.}\label{fig:larock}
\end{figure}

Figure~\ref{fig:larock} instantiates the Larock indole synthesis with ortho-iodoaniline and symmetric 2-butyne to give 2,3-dimethylindole. First follow Pd20. Oxidative addition creates an aryl--palladium bond, into which the alkyne inserts. Now follow N7. Acetate accepts one of its hydrogens as the Pd--N bond forms. Reductive elimination creates the indole C12--N7 bond and regenerates Pd(0) with two phosphine ligands. The tree in Figure~\ref{fig:larock-tree} expands the named reaction into these four events and their electron actions.\cite{kurti,larock}

\subsection{Coverage of two classic named-reaction textbooks}
We use the fourth edition of Jie Jack Li's \emph{Name Reactions} and K\"urti and Czak\'o's \emph{Strategic Applications of Named Reactions in Organic Synthesis}. The page and entry identify a named transformation, while substrate-specific records identify the sequence and level of verification. All 277 main entries in Li are represented. One separately indexed subentry brings the total to 278 records. All 250 chapters of K\"urti and Czak\'o have corresponding entries.\cite{li,kurti} The two catalogues contain 528 entries linked to 321 distinct UMR programs. In the release validation, replaying all 321 programs reproduced their respective recorded outcomes.

Composition accounts for the generality of the representation. Any organic mechanism described by a finite sequence of specified electron, coordination and material changes can be organised using the same state, view and event interfaces. When a substrate requires a new local electronic configuration or event precondition, the relevant rule can be extended with checks specific to that site. The remaining program retains the same composition and replay procedure.

\section{Executing and verifying a mechanism program}
\subsection{Execution, rejection and local repair}
The parser resolves typed objects and references. The executor checks the roles and occupied electron sources, combines the changes from all actions on the same starting state, and checks bond orders, local occupation, formal charges and material balance. The accepted successor becomes available to the next event. The renderer reads those same actions to place the tails, heads and wings of the arrows.

For Figure~\ref{fig:events}a the program submits \code{PAIR(1,2)} and \code{SPLIT(2,3->3)}. The first action consumes the lone pair on N1 and the second assigns the electrons of C2--Br3 to Br3. Deleting the second action returns \texttt{Actions do not match SN2 combination rule} and leaves the starting view in place. This diagnostic identifies the event to revise. A chemist or a language model can keep the bond-forming action, restore the bond-breaking action in the same event and run it again. Execution now yields methylammonium and bromide, and subsequent steps can use this checked state.

\subsection{UMR and Lean as formal languages}
Table~\ref{tab:lean} aligns the inputs, constructed objects and acceptance criteria of the two tasks. Lean tactics can help find a proof; the kernel checks whether the elaborated term has the target type. A model using UMR can explore branches; the executor independently replays its selected path from the initial state. A packaged basic transformation can be reused much as a theorem is reapplied, with the substrate and reagents checked against its preconditions at each instantiation.\cite{lean}

\begin{table}[!htbp]\centering\small\caption{Formal mathematics and mechanism inference as construction-and-checking tasks.}\label{tab:lean}
\begin{tabularx}{\linewidth}{@{}p{.16\linewidth}YY@{}}\toprule
Question & Lean & UMR\\\midrule
Input & A formal proposition, definitions and premises & Starting inventory, Lewis view and conditions, plus the product in explanation tasks\\
Construction & A proof term of the requested type & An atom-mapped event sequence connecting the endpoints\\
Revision & Edit a term or tactic given the current goals and a type error & Edit an event using an electron-source, role or material error and retry from the saved state\\
Acceptance & The term proves the statement under the chosen theory, definitions and axioms & Each event executes under the stated rules and materials, reproducing its successor\\
External judgement & Does the formal statement express the intended problem? & Does the path explain the observations? Could another path also account for them?\\\bottomrule
\end{tabularx}\end{table}

A mathematical proposition may have several proofs, and a chemical transformation may have several paths that account for the electrons and materials involved. In Lean, a term of the target type settles the formal proof obligation. In chemistry, reagents and conditions can favour reaction at one functional group or from one face. Even when each candidate path replays, deciding which one predominates requires evidence about selectivity. We therefore use full reference-pathway agreement to evaluate mechanism inference and also report replay.

\FloatBarrier\Needspace{7\baselineskip}
\section{Mechanism inference by a language model}
\subsection{Mechanism inference with a known product}
We compare two response formats on the \emph{Pathway-only Prediction} task in FukuyamaBench. The model receives starting materials, an observed final product, and ordered conditions, and infers the intervening mechanism. The 319 questions comprise 78 introductory problems (A), 131 graduate-level problems (B) and 110 from research laboratories (C). Both conditions use \texttt{gpt-6.2-astra} with \texttt{medium} reasoning effort, a fresh conversation for each question, and one independent attempt per condition.\cite{fukuyama}

Without UMR, the model responds in the original benchmark format, listing the intermediate species in order as SMILES strings. With UMR and its executor, the model constructs events, inspects the state, admits materials and receives either the successor or an error diagnostic. It can revise an event from a saved intermediate. At submission it selects one branch. The controller independently replays that branch and exports the same per-step product field for the official scorer.

Four few-shot demonstrations matched in chemical content are expressed in the response format of each condition. They illustrate local revision and branch rollback. Starting materials, the target and conditions are visible; reference intermediates and checkpoints are held by the scorer.

The primary metric uses the official ordered structural checkpoints. For each checkpoint, the scorer scans predicted steps after the preceding match and accepts the specified alternative reference steps. The frozen scorer also accepts a nonempty normalized predicted species set that is a subset of an admissible reference set. A question is correct when every required checkpoint is matched in order.

\subsection{Complete reference pathways}
Across all 319 questions, GPT-6.2 Astra matched 62 complete pathways without UMR and 83 with UMR and its executor. The two conditions matched 33 of the same questions. Another 50 were matched only with UMR and 29 only without it, giving a net gain of 21 questions (6.58 percentage points). Appendix~\ref{app:case-scores} reports individual outcomes, including the 13 UMR runs scored zero for missing or unconfirmed submission.

The gain was largest on introductory problems. On A, the count rose from 23/78 to 37/78 (14 more, 17.95 percentage points). On B it rose from 30/131 to 35/131 (five more, 3.82 points), and on C from 9/110 to 11/110 (two more, 1.82 points). Figure~\ref{fig:score-comparison} and Table~\ref{tab:published-pathways} place these results alongside the published benchmark. Our UMR condition exceeds the highest value in the original Table~3 on A, B and C by 2.5, 6.1 and 5.5 points, respectively. The base models and sampling protocols differ across the two studies.\cite{fukuyama}

\begin{center}
\begin{minipage}{.96\linewidth}\centering
\includegraphics[width=\linewidth]{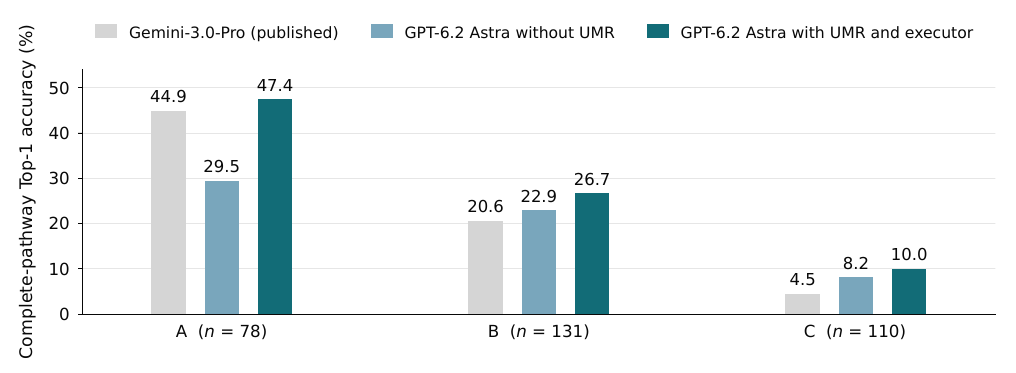}
\captionof{figure}{Complete-pathway Top-1 accuracy on known-product problems. A, B and C denote the introductory, graduate-level and research-laboratory sets, respectively; $n$ is the number of problems in each set. Both GPT-6.2 Astra settings use one independent attempt per problem on the same cases. Gemini-3.0-Pro values are from Table~3 of the FukuyamaBench paper.}\label{fig:score-comparison}
\end{minipage}
\end{center}

\begin{table}[!htbp]\centering\small
\caption{Top-1 accuracy (\%) for complete pathways with a known product. The original paper gives subset scores in its Table~3. Each denominator in this study includes every registered question.}\label{tab:published-pathways}
\begin{tabularx}{\linewidth}{@{}lYrrrr@{}}\toprule
Base model & Inference method & A (78) & B (131) & C (110) & All (319)\\
\midrule
\multicolumn{6}{@{}l}{\itshape Dang et al., published Table 3}\\
GPT-5 & Benchmark protocol & 21.8 & 4.6 & 0.0 & ---\\
Gemini-3.0-Pro & Benchmark protocol & 44.9 & 20.6 & 4.5 & ---\\
Qwen3-235B-A22B & Benchmark protocol & 7.7 & 3.1 & 0.0 & ---\\
Qwen3-30B-A3B & Benchmark protocol & 1.3 & 0.0 & 0.0 & ---\\
ChemDFM-R-14B & Benchmark protocol & 1.3 & 0.0 & 0.0 & ---\\
Authors' post-trained model & Post-training & 10.3 & 1.5 & 0.0 & ---\\
\midrule
\multicolumn{6}{@{}l}{\itshape This study, same base model}\\
GPT-6.2 Astra & Without UMR & 29.5 & 22.9 & 8.2 & 19.4\\
GPT-6.2 Astra & UMR with executor & {\bfseries 47.4} & {\bfseries 26.7} & {\bfseries 10.0} & {\bfseries 26.0}\\
\bottomrule\end{tabularx}\end{table}

The executor replayed all 306 submitted UMR paths, comprising 1,993 events. Of these paths, 83 matched every official checkpoint and 223 missed at least one. Replay checks the electron actions and intermediate structures. The model must still select the reacting site, species and order of events under the given conditions.

Both conditions had the same budget ceilings. Actual usage was 1,229,668 metered tokens without UMR and 83,557,804 with UMR, including cached input (medians of 3,675 and 212,653 per question). The UMR group made 3,602 tool requests. These are results for the complete interactive procedure on one base model. The main analysis uses the frozen evaluator \code{63bb79f9}.

\subsection{Case Study}\label{sec:c035}
\begin{figure}[!htbp]\centering
\includegraphics[width=.94\linewidth]{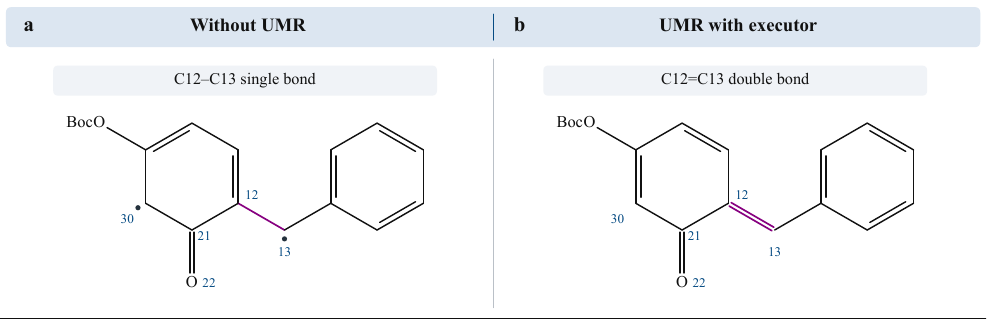}
\caption{C035, submitted intermediate after elimination. The same six-membered ring is shown in both panels. Dots on C13 and C30 give the radical occupations read from the submitted structure. The verified UMR event produces the conjugated, closed-shell intermediate. Boc denotes tert-butoxycarbonyl. Atom numbers follow the original answer.}\label{fig:c035}
\end{figure}

In C035 an elimination generates an \emph{ortho}-quinone methide, which can undergo a hetero-Diels--Alder reaction with ethyl vinyl ether. In step 4 of its answer without UMR, the model wrote that the elimination forms C12=C13. In its submitted structure that bond is single. Reading the submitted SMILES assigns one unpaired electron to C13 and one to C30 (Figure~\ref{fig:c035}). The answer ended at this intermediate without flagging or correcting the bond order. The model also listed unreacted ethyl vinyl ether among the products of earlier steps, which affected species-set matching.

With UMR the model proposed the elimination as one event. Replay gives C12=C13 and C21=O22 in a closed-shell intermediate. During the subsequent cycloaddition it assigned O22 to a role that required carbon. The checker rejected the event with the diagnostic “requires C at mapped atom 22”. This located the mismatch between the atom and its proposed role. Retrying from the preserved intermediate, the model retained the three electron-pair movements and submitted a general pericyclic event that accepts oxygen. This formed O22--C32 and C13--C31 in one step. All five submitted events replay to the archived states.

The official scores were 0/3 and 3/3 checkpoints, respectively. The checkpoints are the alkoxide after Grignard addition, the phenoxide after carbonate transfer and the elimination products, including the quinone methide drawn here. All three precede the final cycloaddition.

\section{Discussion and outlook}
Mechanism inference raises two related questions. The first is whether the structures connect from one step to the next. Given an initial state and a group of electron actions, UMR determines which bonds form or break, where the electrons go and which materials remain available. If an event consumes electrons absent from its starting state or omits an action needed to balance the electron ledger, the executor diagnoses the error at that step. The model can resume from the preceding intermediate.

The second question is selectivity. An intermediate may offer several reactive functional groups, or a reagent may approach the same group from different faces. Each branch can yield a structurally consistent successor. UMR preserves the intermediate and the available materials at the branch point. A model can return to that state, change the site or face of attack, and compare the resulting structures. We plan to use language models capable of assessing selectivity to rank these candidates using the reagents, medium and stereochemical environment, then replay the selected path in UMR. The same process could support forward inference of a mechanism and product, as well as proposing reactants for a given target and checking the proposed mechanism in the forward direction.
\section{Conclusion}
UMR represents polar, radical, pericyclic and organometallic mechanisms with the same elementary-event language. Its verifier and executor check those events and return a specific intermediate from which a model can revise its construction. On 319 known-product problems this combination raised GPT-6.2 Astra's complete-pathway accuracy from 19.4\% to 26.0\%, with 21 additional paths matched overall.

\section*{Code and data availability}
The \href{https://github.com/SrCO3/UMR-Language/releases/tag/v1.0.0}{UMR v1.0.0 release} provides the implementation and runnable examples used to illustrate the language. The software is distributed under the MIT License. The authors' \href{https://github.com/SrCO3/UMR-Language/tree/v1.0.0/named-reactions}{named-reaction catalogue, data and chemical programs} are distributed under CC BY 4.0. The \href{https://github.com/SrCO3/UMR-Language/blob/v1.0.0/NOTICE.md}{release notice} gives the attribution terms.

\section*{Author contributions}
Sirui Shao conceived the study and carried it out. Haichuan Tan provided methodological advice and feedback during the study. Tao Huang and Bowen Gao contributed to research discussions.

\FloatBarrier

\clearpage\appendix
\section{Language, electron actions and element rules}\label{app:source}
A UMR program specifies the material present initially, the Lewis view from which electrons move, and the rules that apply to those atoms. This appendix defines the declarations, action semantics and element rules. Appendix~\ref{app:rules} follows those declarations through a single event and a mechanism.

\subsection{States, views and events}
UMR accepts a keyword-based \code{.umr} document or JSON conforming to the accompanying schema; both parse to the same typed abstract syntax tree. A document specifies its initial state and primary Lewis view, followed by events originating from known views. Atom identifiers retain their identity in leaving groups, byproducts and later steps. A \code{state_id} identifies a chemical state; a \code{view_id} identifies one Lewis representation of that state. Each event refers to a reactant view and names its successor. Alternative branches may share a reactant view.

\begin{center}\small
\begin{tabularx}{\linewidth}{@{}p{28mm}YY@{}}\toprule
Declaration&Information supplied&Execution meaning\\\midrule
\code{STATE}, \code{VIEW}&State and primary view; mapped atoms, bonds, formal charges and local occupations&Establish the initial inventory; retain explicitly unknown electronic states\\
\code{REACT}&Reactant view, successor identifiers, event type and required roles; authored actions or a registered expansion&Read one reactant view and commit one checked successor\\
\code{RESONANCE}&Source and target views of one state, an allowed rule and its actions&Check another Lewis representation without adding a reaction step\\
\code{PHASE}, \code{REAGENT}&Stage, activated atoms and reagent identity/roles&Constrain the materials and conditions available in a stage\\
\code{CYCLE}&Executed events, carrier mapping, regenerated products and persistent core&Check closure separately from the constituent events\\
\code{VALENCE}, \code{INTERMEDIATE}&Local declarations bound to a view and site&Check the declared local scope; a name does not supply electrons\\
\bottomrule\end{tabularx}
\end{center}

The \code{hydrogens} field counts attached, unnumbered hydrogens; a transferred hydrogen requires its own \code{atom_id}. The \code{nonbonding_electrons} count includes paired and unpaired local electrons, with the latter additionally specified by \code{radical_electrons}. Metal oxidation states, ligand assignments and coordination bindings have separate fields and are not replaced by atomic formal charge. Missing stereochemical relationships remain unknown; a two-dimensional drawing does not establish absolute configuration.

Section~\ref{app:res} shows how one state can have two views. \code{STATE.primary_view_id} points to the primary view, and \code{VIEW.state_id} points back to its state. Within \code{REACT}, \code{reactant_view_id} selects the particular drawing on which the arrows act, while \code{product_state_id} and \code{product_view_id} name the two objects created on success. A later step refers to a previously reached view. Atom identities persist along the path.

\subsection{Resonance structures}\label{app:resonance}
A \code{RESONANCE} transformation changes the view of a single $S$. In the left contributor of Figure~\ref{fig:enolate-views}, the enolate bears its formal negative charge on C1. Assigning the C1 lone pair to the C1--C2 bond and a pair of C2=O3 $\pi$ electrons to O3 gives the contributor on the right. The two curved arrows describe how the drawings represent the same delocalized electronic structure. The checker verifies atom identities, $\sigma$ connectivity, hydrogen attachments, net charge, electronic state and applicable stereochemical relationships. Both views belong to \code{s0}. Replay records one change of representation and zero reaction steps. Moving a hydrogen to another atom to form a tautomer changes hydrogen connectivity and requires a reaction event.

\par\medskip\noindent\begin{minipage}{\linewidth}\centering
\includegraphics[width=.78\linewidth]{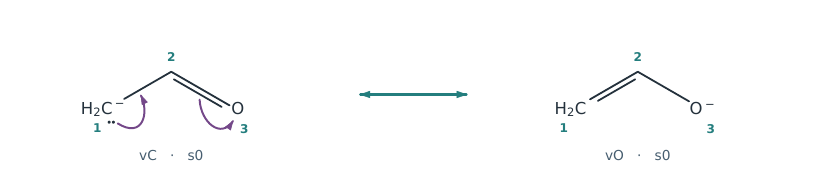}
\captionof{figure}{Two Lewis views of one enolate. The double-headed arrow connects the carbon-centred contributor \code{vC} and the oxygen-centred contributor \code{vO}. Both retain the C1--C2--O3 skeleton, the hydrogen attachments and a net charge of $-1$. The native program submits \code{PAIR(1,2)} and \code{SPLIT(2,3->3)} together.}\label{fig:enolate-views}
\end{minipage}\par\medskip

\subsection{Declaration grammar}
The following grammar covers the top-level declarations recognised by the parser. Angle brackets denote nonterminals, square brackets an option, an asterisk repetition and quotes literal punctuation. Bodies are JSON objects. If a \code{REACT} declaration omits its action block, its registered rule determines execution; \code{RESONANCE} requires an action block.
\begin{Verbatim}[fontsize=\footnotesize]
<document> := UMR 1.0 INITIAL <id> ";" <declaration>*
<declaration> := (RECORD | META) <object> ";"
               | <kind> <id> <object> ";"
               | REACT <id> <object> [<block>] ";"
               | RESONANCE <id> <object> <block> ";"
<kind> := STATE | VIEW | PHASE | REAGENT | CYCLE | VALENCE
        | INTERMEDIATE | ANALYZE | EXCEPTION | OBSERVE
<block> := DO "{" (<action> ";")* "}"
\end{Verbatim}
\code{RECORD} carries record identity, revision and symbolic mode; \code{META} carries metadata. \code{ANALYZE} registers an analysis request, \code{EXCEPTION} associates evidence with a scoped exception, and \code{OBSERVE} declares an observation of a type supported by this version. Semicolons terminate declarations. A hash character starts a line comment. The accompanying \code{schema/umr-1.0.json} provides the complete field types, defaults and enumerations.

In the text format, the declaration header supplies the identifier and a \code{DO} block supplies the arrows. The parser transfers identifiers into typed fields such as \code{state_id}, \code{view_id} and \code{event_id}. For example, \code{REACT e1} supplies \code{event_id="e1"}. Repeating that field in the body yields a parse error before any action executes. The JSON form supplies the fields and action arrays directly.

\subsection{Local semantics of electron actions}
The four actions specify the destinations of electron pairs or individual electrons. Bond changes in the table are aggregated within a complete event; an action alone is generally not an elementary reaction. The substitution, radical substitution, pericyclic event and oxidative addition in Fig.~\ref{fig:events} illustrate their combinations.

\begin{center}\small
\begin{tabularx}{\linewidth}{@{}p{37mm}YY@{}}\toprule
Action&Local change&Resource check\\\midrule
\code{PAIR(a,b)}&Take a lone pair from $a$ and increase the $a$--$b$ bond order by one&The reactant has an available pair; coordination-reserved electrons cannot be spent twice\\
\code{SPLIT(a,b->b)}&Decrease the $a$--$b$ bond order by one and assign that pair to $b$&The source bond pair exists; the destination must be the second atom, with arguments reversed for the opposite direction\\
\code{SHIFT(a,b,c)}&Decrease $a$--$b$ by one and increase $b$--$c$ by one&The consumed pair exists in the reactant view\\
\code{FISHHOOK(src,dst)}&Move one indexed electron to an atom or a bond pool&The source slot is occupied and used once; both slots of a consumed bond pair need destinations; a new ordinary bond pool receives two electrons\\
\bottomrule\end{tabularx}
\end{center}

For the three pair instructions, write $b_{ij}$ for bond order and $n_i$ for the number of nonbonding electrons on atom $i$. \code{PAIR(a,b)} contributes $(\Delta n_a,\Delta b_{ab})=(-2,+1)$, \code{SPLIT(a,b->b)} contributes $(\Delta b_{ab},\Delta n_b)=(-1,+2)$, and \code{SHIFT(a,b,c)} contributes $(\Delta b_{ab},\Delta b_{bc})=(-1,+1)$. The executor checks all electron sources in the reactant view before summing these changes. Two arrows can therefore complete a substitution with one successor.

\code{FISHHOOK} is a single-electron representation used in radical bond changes and supported electron-transfer events. \code{RAD(a)} selects an unpaired electron; \code{BOND(a,b,k)} selects slot zero or one of a $\sigma$ bond; \code{NEW(a,b)} and \code{ATOM(a)} denote a bond pool and an atomic destination. Full JSON endpoints additionally specify $\pi$ bond pairs, other radical slots and individual electrons in lone pairs.

\subsection{A worked UMR example}\label{app:substitution-program}
The following input reproduces the nucleophilic substitution in Fig.~\ref{fig:events}a. \code{PAIR} draws two electrons from the existing lone pair on N1; \code{SPLIT} assigns the C2--Br3 bonding pair to Br3. Both instructions execute in one event.
\begin{Verbatim}[fontsize=\footnotesize,samepage=true]
UMR 1.0 INITIAL s0;
STATE s0 {"primary_view_id":"v0","total_charge":0};
VIEW v0 {"state_id":"s0","atoms":[
  {"atom_id":1,"element":"N","formal_charge":0,
   "nonbonding_electrons":2,"hydrogens":3},
  {"atom_id":2,"element":"C","formal_charge":0,
   "nonbonding_electrons":0,"hydrogens":3},
  {"atom_id":3,"element":"Br","formal_charge":0,
   "nonbonding_electrons":6}],
  "bonds":[{"a":2,"b":3,"order":1}]};
REACT e1 {"event_type":"SN2","reactant_view_id":"v0",
  "product_state_id":"s1","product_view_id":"v1",
  "bindings":{"nucleophile":1,"electrophile":2,
              "leaving_group":3}} DO {
  PAIR(1,2);
  SPLIT(2,3->3);
};
\end{Verbatim}

From the root of the \href{https://github.com/SrCO3/UMR-Language/releases/tag/v1.0.0}{UMR v1.0.0 release}, run
\begin{Verbatim}[fontsize=\small]
python umr.py examples/paper-1.0/substitution.umr
python umr.py examples/paper-1.0/substitution-omitted-arrow.umr
\end{Verbatim}
The first command produces \code{s1/v1}: N1--C2 forms, C2--Br3 breaks, and the formal charges on N1 and Br3 become $+1$ and $-1$, respectively, while the total charge remains zero. Omitting \code{SPLIT} returns \texttt{Actions do not match SN2 combination rule} and commits no successor. This rejection concerns a mismatch between the declared event and its action combination.

\subsection{Element rules and electron counting}\label{app:element-rules}
UMR records the identities of all 118 elements and supplies electron-counting rules for 67. Figure~\ref{fig:umr-element-scope} shows the two validation routes. For 19 elements, local main-group rules check electron occupation and bonding in a Lewis view. For 48 elements, coordination rules check how electrons are assigned to a centre and its ligands. Within a complex, both the local electron occupation of a ligand atom and the coordination electron count of the metal centre can be checked. Grey cells indicate elements whose identity is retained and for which local electronic changes still lack a built-in rule.

\begin{center}\begin{minipage}{\linewidth}\centering
\includegraphics[width=.95\linewidth]{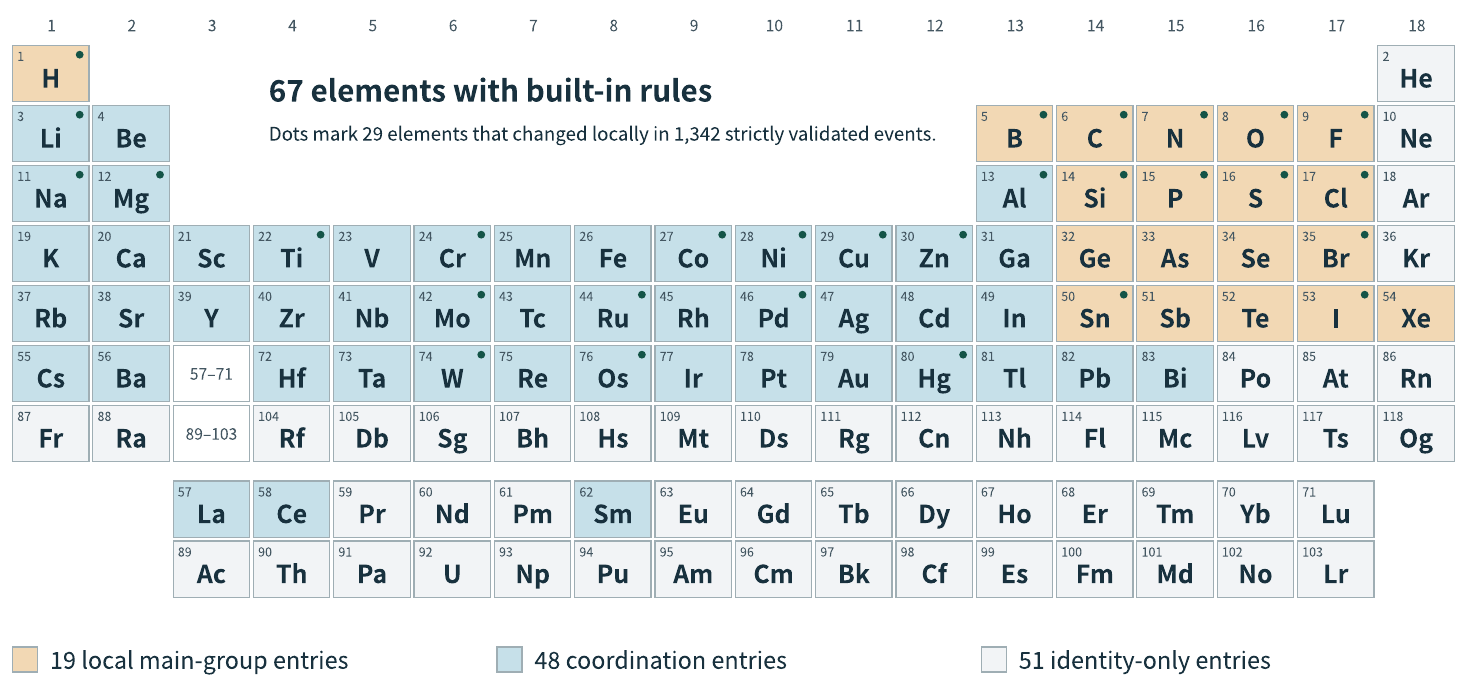}
\captionof{figure}{Element rules in UMR. Colours distinguish local main-group rules, coordination rules and elements retained by identity alone. Dots mark elements whose local bonding or electronic state changed in the replayed mechanism archive.}\label{fig:umr-element-scope}
\end{minipage}\end{center}

\paragraph{Local main-group rules.} UMR uses bond orders, nonbonding electrons and unpaired electrons to check formal charge, then applies element-specific constraints on coordination and shell occupation. The table below gives checked oxidation states and selected coordination constraints. Coordination number counts directly attached atoms, so a doubly bonded neighbour counts once. Table~\ref{tab:element-registry} gives the complete register or the applicable basis for local validation.

\begingroup\small
\setlength{\tabcolsep}{3.5pt}
\renewcommand{\LTcaptype}{}
\begin{longtable}{@{}>{\raggedright\arraybackslash}p{.135\linewidth}>{\raggedright\arraybackslash}p{.245\linewidth}>{\raggedright\arraybackslash}p{\dimexpr.62\linewidth-4\tabcolsep\relax}@{}}
\toprule
Elements & Checked oxidation states & Coordination and local checks\\\midrule\endfirsthead
\toprule
Elements & Checked oxidation states & Coordination and local checks\\\midrule\endhead
\midrule\endfoot
\bottomrule\endlastfoot
H & $-1,+1$ & Coordination number 0 when free and 1 when singly bonded. The shell is full at two electrons.\\
B & $+3$ & A neutral three-coordinate centre has six shell electrons. A four-coordinate centre with formal charge $-1$ has eight.\\
C & $-4$ to $+4$ & At a centre with three single bonds, nonbonding occupation distinguishes a cation, radical and anion.\\
Si, Sn & $-4,+2,+4$ & Low-coordinate centres require a compatible lone-pair occupation and spin declaration. Higher coordination also depends on the neighbours and bond orders.\\
Ge & $-4$ to $+4$ & Bonding and nonbonding electrons are checked locally. Coordination numbers 4, 5 and 6 can share an oxidation state.\\
N & $-3,-2,-1$ and $+1$ to $+5$ & Bond orders, formal charge and unpaired electrons determine the local closed-shell or radical configuration.\\
P & $-3,+3,+5$ & Both coordination number and bond-order sum are checked. Six-coordinate P(V) requires six terminal F or six terminal Cl ligands.\\
As, Sb & $-3$ to $+5$ & Neighbours, lone pairs and shell occupation distinguish five- and six-coordinate centres in oxidation state $+5$.\\
O & $-2,-1$ & Bond orders and lone pairs determine the formal charge at the specified coordination.\\
S, Se, Te & $-2,+2,+4,+6$ (Te also admits the intervening integers) & S and Se admit 8, 10 or 12 shell electrons in the matching configuration. Four- and six-coordinate Te also require the appropriate bond orders.\\
F & $-1$ & The free anion and singly bonded states are checked for an octet.\\
Cl, Br, I & $-1,+1,+3,+5,+7$ & Expanded shells depend on the ligands. The general expanded-shell rule for Cl and Br requires O or F neighbours and no implicit H on the centre.\\
Xe & $0,+2,+4,+6,+8$ & Two-, four- and six-coordinate centres are distinguished by bond orders and lone-pair counts.\\
\end{longtable}\endgroup

\paragraph{Coordination rules.} Ligand binding sites determine the coordination number. For a d-block centre under ionic counting, the group number minus the oxidation state gives the formal d count. Adding the electrons donated by ligands gives the valence-electron count at the centre. For s- and p-block centres, ionic valence-electron counting is available and the d count is inapplicable. La, Ce and Sm also use the coordination route. For example, in $\mathrm{Cr(CO)_6}$, Cr(0) is d$^6$ and six neutral CO ligands each donate two electrons, giving 18 valence electrons.

In $\mathrm{Pd(PPh_3)_2}$, the starting complex in Figure~\ref{fig:larock}, two neutral triphenylphosphine ligands bind Pd20 through P50 and P51. Each donates two electrons. Under ionic counting, Pd(0) is d$^{10}$, with coordination number 2 and 14 valence electrons. Local Lewis rules describe pair donation to a main-group centre. In $\mathrm{AsF_5}+\mathrm{F}^{-}\to[\mathrm{AsF_6}]^{-}$, \code{PAIR(7,1)} uses a lone pair on F7 to form the F7--As1 bond. The As coordination number rises from 5 to 6, its shell count from 10 to 12 and its formal charge changes from 0 to $-1$, while its oxidation state remains $+\mathrm{V}$.

\subsection{Oxidation states by element}\label{app:element-os}
Table~\ref{tab:element-registry} lists all 67 elements with built-in rules in atomic-number order. Common oxidation states are the union of those listed in the \href{https://periodic-table.rsc.org/}{Royal Society of Chemistry periodic table}, the \href{https://pubchem.ncbi.nlm.nih.gov/rest/pug/periodictable/JSON}{PubChem} \code{OxidationStates} field and the \href{https://pypi.org/project/mendeleev/1.1.0/}{mendeleev} \code{main} field. The last column gives the discrete registered states or the applicable local configuration rule. Both the initial and the resulting structures must satisfy the corresponding bonding and occupation constraints.

\begingroup\small
\setlength{\tabcolsep}{3.5pt}\renewcommand{\arraystretch}{1.10}
\setlength{\LTpre}{.6\baselineskip}\setlength{\LTpost}{.6\baselineskip}
\begin{longtable}{@{}rll>{\raggedright\arraybackslash}p{.32\linewidth}>{\raggedright\arraybackslash}p{.465\linewidth}@{}}
\caption{The 67 elements with built-in rules and their oxidation states, in atomic-number order.}\label{tab:element-registry} \\
\toprule
\textbf{$Z$} & \textbf{Atom} & \textbf{Rule} & \textbf{Common oxidation states} & \textbf{Registered states or local check} \\
\midrule\endfirsthead
\multicolumn{5}{@{}l}{\small\itshape Element rules and oxidation states (continued)} \\
\toprule
\textbf{$Z$} & \textbf{Atom} & \textbf{Rule} & \textbf{Common oxidation states} & \textbf{Registered states or local check} \\
\midrule\endhead
\midrule\multicolumn{5}{r@{}}{\small\itshape Continued on next page} \\
\endfoot
\bottomrule\endlastfoot
1 & H & Local & $-1$, $+1$ & \emph{local profile} \\
3 & Li & Coord. & $+1$ & $0$, $+1$ \\
4 & Be & Coord. & $+2$ & $0$, $+1$, $+2$ \\
5 & B & Local & $+3$ & \emph{local profile} \\
6 & C & Local & $-4$, $-3$, $-2$, $-1$, $0$, $+1$, $+2$, $+3$, $+4$ & \emph{local profile} \\
7 & N & Local & $-3$, $-2$, $-1$, $+1$, $+2$, $+3$, $+4$, $+5$ & \emph{local profile} \\
8 & O & Local & $-2$, $-1$ & \emph{local profile} \\
9 & F & Local & $-1$ & \emph{local profile} \\
11 & Na & Coord. & $+1$ & $-1$, $0$, $+1$ \\
12 & Mg & Coord. & $+2$ & $0$, $+1$, $+2$ \\
13 & Al & Coord. & $+3$ & $-2$, $-1$, $0$, $+1$, $+2$, $+3$ \\
14 & Si & Local & $-4$, $+2$, $+4$ & \emph{local profile} \\
15 & P & Local & $-3$, $+3$, $+5$ & \emph{local profile} \\
16 & S & Local & $-2$, $+2$, $+4$, $+6$ & \emph{local profile} \\
17 & Cl & Local & $-1$, $+1$, $+3$, $+5$, $+7$ & \emph{local profile} \\
19 & K & Coord. & $+1$ & $-1$, $0$, $+1$ \\
20 & Ca & Coord. & $+2$ & $0$, $+1$, $+2$ \\
21 & Sc & Coord. & $+3$ & $0$, $+1$, $+2$, $+3$ \\
22 & Ti & Coord. & $+2$, $+3$, $+4$ & $-2$, $-1$, $0$, $+1$, $+2$, $+3$, $+4$ \\
23 & V & Coord. & $0$, $+2$, $+3$, $+4$, $+5$ & $-3$, $-1$, $0$, $+1$, $+2$, $+3$, $+4$, $+5$ \\
24 & Cr & Coord. & $0$, $+2$, $+3$, $+6$ & $-4$, $-2$, $-1$, $0$, $+1$, $+2$, $+3$, $+4$, $+5$, $+6$ \\
25 & Mn & Coord. & $-1$, $0$, $+2$, $+3$, $+4$, $+6$, $+7$ & $-3$, $-1$, $0$, $+1$, $+2$, $+3$, $+4$, $+5$, $+6$, $+7$ \\
26 & Fe & Coord. & $-2$, $0$, $+2$, $+3$, $+6$ & $-4$, $-2$, $-1$, $0$, $+1$, $+2$, $+3$, $+4$, $+5$, $+6$, $+7$ \\
27 & Co & Coord. & $-1$, $0$, $+2$, $+3$ & $-3$, $-1$, $0$, $+1$, $+2$, $+3$, $+4$, $+5$ \\
28 & Ni & Coord. & $0$, $+2$, $+3$ & $-2$, $-1$, $0$, $+1$, $+2$, $+3$, $+4$ \\
29 & Cu & Coord. & $+1$, $+2$ & $-2$, $0$, $+1$, $+2$, $+3$, $+4$ \\
30 & Zn & Coord. & $+2$ & $-2$, $0$, $+1$, $+2$ \\
31 & Ga & Coord. & $+3$ & $-5$, $-4$, $-3$, $-2$, $-1$, $0$, $+1$, $+2$, $+3$ \\
32 & Ge & Local & $-4$, $+2$, $+4$ & $-4$, $-3$, $-2$, $-1$, $0$, $+1$, $+2$, $+3$, $+4$ \\
33 & As & Local & $-3$, $+3$, $+5$ & $-3$, $-2$, $-1$, $0$, $+1$, $+2$, $+3$, $+4$, $+5$ \\
34 & Se & Local & $-2$, $+2$, $+4$, $+6$ & \emph{local profile} \\
35 & Br & Local & $-1$, $+1$, $+3$, $+5$, $+7$ & \emph{local profile} \\
37 & Rb & Coord. & $+1$ & $-1$, $0$, $+1$ \\
38 & Sr & Coord. & $+2$ & $0$, $+1$, $+2$ \\
39 & Y & Coord. & $+3$ & $0$, $+1$, $+2$, $+3$ \\
40 & Zr & Coord. & $+4$ & $0$, $+1$, $+2$, $+3$, $+4$ \\
41 & Nb & Coord. & $+3$, $+5$ & $-3$, $-1$, $0$, $+1$, $+2$, $+3$, $+4$, $+5$ \\
42 & Mo & Coord. & $0$, $+2$, $+3$, $+4$, $+5$, $+6$ & $-4$, $-2$, $-1$, $0$, $+1$, $+2$, $+3$, $+4$, $+5$, $+6$ \\
43 & Tc & Coord. & $+4$, $+6$, $+7$ & $-3$, $-1$, $0$, $+1$, $+2$, $+3$, $+4$, $+5$, $+6$, $+7$ \\
44 & Ru & Coord. & $-2$, $0$, $+2$, $+3$, $+4$, $+6$, $+8$ & $-4$, $-2$, $0$, $+1$, $+2$, $+3$, $+4$, $+5$, $+6$, $+7$, $+8$ \\
45 & Rh & Coord. & $0$, $+1$, $+2$, $+3$, $+4$, $+5$ & $-3$, $-1$, $0$, $+1$, $+2$, $+3$, $+4$, $+5$, $+6$, $+7$ \\
46 & Pd & Coord. & $0$, $+2$, $+3$, $+4$ & $0$, $+1$, $+2$, $+3$, $+4$, $+5$ \\
47 & Ag & Coord. & $+1$, $+2$ & $-2$, $-1$, $0$, $+1$, $+2$, $+3$ \\
48 & Cd & Coord. & $+2$ & $-2$, $0$, $+1$, $+2$ \\
49 & In & Coord. & $+3$ & $-5$, $-2$, $-1$, $0$, $+1$, $+2$, $+3$ \\
50 & Sn & Local & $-4$, $+2$, $+4$ & \emph{local profile} \\
51 & Sb & Local & $-3$, $+3$, $+5$ & $-3$, $-2$, $-1$, $0$, $+1$, $+2$, $+3$, $+4$, $+5$ \\
52 & Te & Local & $-2$, $+2$, $+4$, $+6$ & $-2$, $-1$, $0$, $+1$, $+2$, $+3$, $+4$, $+5$, $+6$ \\
53 & I & Local & $-1$, $+1$, $+3$, $+5$, $+7$ & \emph{local profile} \\
54 & Xe & Local & $0$, $+2$, $+4$, $+6$ & $0$, $+2$, $+4$, $+6$, $+8$ \\
55 & Cs & Coord. & $+1$ & $-1$, $0$, $+1$ \\
56 & Ba & Coord. & $+2$ & $0$, $+1$, $+2$ \\
57 & La & Coord. & $+3$ & $+3$ \\
58 & Ce & Coord. & $+3$, $+4$ & $+2$, $+3$, $+4$ \\
62 & Sm & Coord. & $+2$, $+3$ & $0$, $+2$, $+3$ \\
72 & Hf & Coord. & $+4$ & $-2$, $0$, $+1$, $+2$, $+3$, $+4$ \\
73 & Ta & Coord. & $+5$ & $-3$, $-1$, $0$, $+1$, $+2$, $+3$, $+4$, $+5$ \\
74 & W & Coord. & $0$, $+2$, $+3$, $+4$, $+5$, $+6$ & $-4$, $-2$, $-1$, $0$, $+1$, $+2$, $+3$, $+4$, $+5$, $+6$ \\
75 & Re & Coord. & $-1$, $+2$, $+4$, $+6$, $+7$ & $-3$, $-1$, $0$, $+1$, $+2$, $+3$, $+4$, $+5$, $+6$, $+7$ \\
76 & Os & Coord. & $-2$, $0$, $+2$, $+3$, $+4$, $+6$, $+8$ & $-4$, $-2$, $-1$, $0$, $+1$, $+2$, $+3$, $+4$, $+5$, $+6$, $+7$, $+8$ \\
77 & Ir & Coord. & $-1$, $0$, $+1$, $+2$, $+3$, $+4$, $+6$ & $-3$, $-2$, $-1$, $0$, $+1$, $+2$, $+3$, $+4$, $+5$, $+6$, $+7$, $+8$, $+9$ \\
78 & Pt & Coord. & $+2$, $+4$ & $-3$, $-2$, $-1$, $0$, $+1$, $+2$, $+3$, $+4$, $+5$, $+6$ \\
79 & Au & Coord. & $-1$, $+1$, $+2$, $+3$, $+4$, $+5$ & $-3$, $-2$, $-1$, $0$, $+1$, $+2$, $+3$, $+4$, $+5$ \\
80 & Hg & Coord. & $+1$, $+2$ & $-2$, $0$, $+1$, $+2$ \\
81 & Tl & Coord. & $+1$, $+3$ & $-5$, $-2$, $-1$, $0$, $+1$, $+2$, $+3$ \\
82 & Pb & Coord. & $+2$, $+4$ & $-4$, $-2$, $-1$, $0$, $+1$, $+2$, $+3$, $+4$ \\
83 & Bi & Coord. & $+3$, $+5$ & $-3$, $-2$, $-1$, $0$, $+1$, $+2$, $+3$, $+4$, $+5$ \\
\end{longtable}\endgroup

These rules cover the oxidation states of main-group elements, metal reagents and catalysts commonly encountered in organic reactions. A reagent that requires a different valence or coordination environment can be accommodated by extending the element rule with its oxidation state, coordination and electron occupation, then checking the relevant steps under that rule.

\needspace{8\baselineskip}\section{Event validation and composition of mechanism programs}\label{app:rules}
Appendix~\ref{app:source} defined the objects in an instruction. To execute an event, the checker verifies that the reactant view supplies the electrons for every action and that the resulting state is available to the next event. Figure~\ref{fig:larock-tree} identifies these levels in the path drawn in Fig.~\ref{fig:larock}.

\subsection{From role bindings to event checks}
Roles give persistent atom identifiers their meaning in an event. In the preceding example, \code{bindings} assigns the nucleophile, electrophile and leaving group to atoms 1, 2 and 3. Interchanging two values changes the proposed event. The selected rule checks the roles, existing bonds, required electrons and full action group. A registered template can expand suitable roles into actions. Authored arrows are subject to the corresponding combination rule.

\begin{center}\small
\begin{tabularx}{\linewidth}{@{}p{30mm}YY@{}}\toprule
Event family&Required declarations or matches&Checking scope\\\midrule
Polar bond changes and addition&Nucleophilic/electrophilic sites, existing bonds and the selected polar rule&Electron sources, local bond orders and valence; carbonyl rules check C/O roles\\
Proton transfer&Base, mapped H and proton donor&Breaking and forming bonds to the same H; changing an implicit hydrogen count is insufficient\\
Substitution and elimination&Nucleophile/base, leaving group, reaction center and any transferred H&Action combination and reactant structure; mapped SN2 inversion when defined, with E2 geometry reported from supplied evidence\\
Radical and single-electron processes&Electron slots, donor/acceptor, local occupation and electronic-state declarations&Slot and bond-pool accounting, total ledger and supported electronic states; radical counts do not determine the physical ground state\\
Pericyclic processes&Registered template, ordered roles, reactive region and thermal/photochemical declarations&Template topology, concerted actions and implemented orbital/stereochemical constraints\\
Coordination and organometallic processes&Metal, ligands, binding modes and the corresponding event rule&Coordination, oxidation state and electron counts within scope; donor-owned coordination and covalent electron allocation remain distinct\\
\bottomrule\end{tabularx}
\end{center}

Common checks also cover atom references and identities, available electrons, bond-order ranges, consistency of formal charge and occupation, hydrogen and material inventories, total charge, and declared stages and dependencies. Element rules, local declarations and dedicated coordination models determine the checking scope. An unsupported model is reported as such rather than replaced by a universal octet or eighteen-electron rule. Stereochemical checks preserve mapped local relationships and recalculate applicable CIP labels.

\subsection{Successors, diagnostics and path replay}
For the substitution in Section~\ref{app:substitution-program}, the successful receipt for \code{e1} connects \code{v0} to \code{v1}. The nonbonding count falls from two to zero on N1 and rises from six to eight on Br3. The bond-order sum on C2 remains four. Taken together, the two actions conserve electrons, atom identity and total charge. Removing only the bond-cleavage arrow triggers the action-combination check. The result retains one initial state, one view and zero executed events. The event identifier locates the proposal to revise.

Replay executes the submitted events again from the declared initial state. The receipt field \code{execution_complete} records completion of the declared workflow and \code{full_electron_path_verified} records formal verification of the full electron path. Net transformations, observed endpoints, manually supplied bridges and applied exceptions retain their evidence status when later steps refer to the resulting states. The field \code{stereochemistry_fully_specified} records stereochemical completeness separately. The external evaluator compares ordered reference checkpoints in the model experiment.

\subsection{Stages, branches and cycle closure}
A \code{PHASE} identifies its events through \code{event_ids} and the mapped atoms activated at that stage through \code{activate_atom_ids}. This permits later addition of a registered reagent. Different events can read the same reactant view and each successful event names its own successor. Consecutive source and target identifiers define the path to replay once a branch is chosen.

For a declared \code{CYCLE}, the checker follows \code{event_ids} from the initial to the final view to establish continuity. It uses \code{carrier_mapping} to compare carrier atoms, bonds and coordination at the two ends. It checks the persistent core, product release and overall material and electron balance. A \code{closed_one_turnover} receipt records one executed and closed turnover. Additional turnovers require their own records.

\needspace{.62\textheight}\subsection{Expanding a named organic reaction into a tree}\label{sec:larock-tree}
Following a named reaction down to one curved arrow takes two refinements. First, decompose the Larock indole synthesis into four basic transformations. Then bind the chemical roles in each transformation to the atoms of this substrate to obtain executable events. Figure~\ref{fig:larock-tree} expands the teaching instance \code{STR-260}, indexed to p.~260 of K\"urti and Czak\'o. Figure~\ref{fig:larock} shows the structures and arrows for the same instance. The initial state contains preformed $\mathrm{Pd(PPh_3)_2}$ and symmetric 2-butyne.\cite{kurti,larock}

\par\smallskip\noindent\begin{minipage}{\linewidth}\centering
\includegraphics[width=\linewidth]{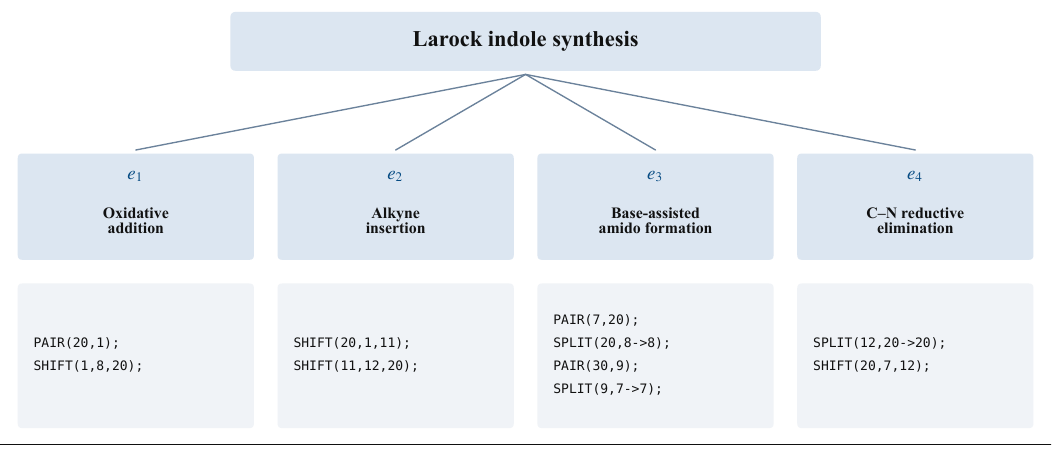}
\captionof{figure}{One named reaction expands directly into four events, ordered from left to right as $e_1$ through $e_4$. The ten native instructions below their respective events correspond to the arrows in Figure~\ref{fig:larock}. All actions in one event read the same pre-state and jointly produce one successor.}\label{fig:larock-tree}
\end{minipage}\par\smallskip

In $e_3$, the lone pair on N7 forms the Pd--N bond and I8 receives the electrons of the Pd20--I8 bond. At the same time acetate O30 accepts H9, and the N7--H9 bond electrons return to N7. The executor commits all four actions together to give $s_3$. Event $e_4$ then forms C12--N7, releasing indole and regenerating Pd(0) with its two phosphine ligands. Event $e_2$ groups alkyne association and migratory insertion between the recorded arylpalladium and vinylpalladium endpoints. The four transitions replay in sequence from $s_0$ to $s_4$ in \code{examples/paper-1.0/larock.umr}.

\clearpage\section{Per-case official benchmark scores}
\label{app:case-scores}

Tables~\ref{tab:case-scores-A}--\ref{tab:case-scores-C} report all 319 registered questions in the known-product pathway task, with one attempt per method and question. D denotes \textbf{Without UMR}; U denotes \textbf{UMR with executor}. For each method, $s\in\{0,1\}$ is the archived official complete-path score, and $m/n$ gives the number of matched reference checkpoints over the number required for that question. The scorer checks reference checkpoints in order. A score of 1 denotes a complete reference-path match under that scorer; it does not independently validate the electron actions, reaction mechanism, or every generated intermediate. The partial count $m$ is reported separately and is not an additional complete path.

All 319 D positions and 306 U positions contain submissions. The remaining 13 U positions are retained in the denominator and distinguished in the tables: $\dagger$ marks 11 positions with no native submission, and $\ddagger$ marks two positions for which interrupted delivery leaves submission status unknown. Their frozen official scores are 0 with 0 matched checkpoints, following the registered failure-accounting rule. These recorded zeros do not establish that a chemical path was generated and found incorrect. Unmarked U positions have submitted paths.

Table~\ref{tab:case-scores-totals} cross-checks the case-level records. Required checkpoints total 1,198; the two methods match 367 and 434 of these checkpoints, respectively. Complete-path totals are 62/319 for D and 83/319 for U. Within each partition, read the left panel from top to bottom, followed by the right panel from top to bottom; headers repeat across page breaks. Every case appears exactly once in the case tables.

\begingroup
\small
\begin{table}[htbp]
\centering
\caption{Cross-check of partition subtotals and overall totals}\label{tab:case-scores-totals}
\setlength{\tabcolsep}{4.4pt}
\begin{tabular}{@{}lrrrrrrrr@{}}
\toprule
Set & Cases & Required & D full & D matched & U full & U matched & U $\dagger$ & U $\ddagger$ \\
\midrule
A & 78 & 252 & 23 & 104 & 37 & 133 & 2 & 0 \\
B & 131 & 453 & 30 & 137 & 35 & 173 & 5 & 0 \\
C & 110 & 493 & 9 & 126 & 11 & 128 & 4 & 2 \\
\midrule
All & 319 & 1198 & 62 & 367 & 83 & 434 & 11 & 2 \\
\bottomrule
\end{tabular}
\end{table}

\endgroup
\begingroup
\fontsize{10}{12}\selectfont
\setlength{\tabcolsep}{6pt}
\renewcommand{\arraystretch}{1.12}
\setlength{\LTpre}{0.7\baselineskip}
\setlength{\LTpost}{0.8\baselineskip}
\begin{longtable}{@{}lcccc@{\hspace{2.4em}}lcccc@{}}
\caption{Official per-case scores: partition A (78 cases)}\label{tab:case-scores-A}\\
\toprule
\textbf{Case} & \multicolumn{2}{c}{\textbf{D}} & \multicolumn{2}{c}{\textbf{U}} & \textbf{Case} & \multicolumn{2}{c}{\textbf{D}} & \multicolumn{2}{c}{\textbf{U}} \\
 & $s$ & $m/n$ & $s$ & $m/n$ & & $s$ & $m/n$ & $s$ & $m/n$ \\
\midrule
\endfirsthead
\multicolumn{10}{@{}l}{\small\itshape Partition A (continued)}\\
\toprule
\textbf{Case} & \multicolumn{2}{c}{\textbf{D}} & \multicolumn{2}{c}{\textbf{U}} & \textbf{Case} & \multicolumn{2}{c}{\textbf{D}} & \multicolumn{2}{c}{\textbf{U}} \\
 & $s$ & $m/n$ & $s$ & $m/n$ & & $s$ & $m/n$ & $s$ & $m/n$ \\
\midrule
\endhead
\midrule\multicolumn{10}{r@{}}{\small\itshape Continued on next page}\\
\endfoot
\bottomrule
\endlastfoot
A001 & 1 & 2/2 & 1 & 2/2 & A040 & 0 & 0/1 & 0 & 0/1 \\
A002 & 0 & 0/3 & 1 & 3/3 & A041 & 1 & 1/1 & 1 & 1/1 \\
A003 & 0 & 0/3 & 0 & 0/3 & A042 & 0 & 0/1 & 1 & 1/1 \\
A004 & 0 & 0/3 & 0 & 1/3 & A043 & 0 & 0/4 & 0 & 2/4 \\
A005 & 1 & 3/3 & 0 & 0/3 & A044 & 0 & 1/2 & 0 & 1/2 \\
A006 & 1 & 2/2 & 0 & 0/2 & A045 & 0 & 0/2 & 0 & 1/2 \\
A007 & 0 & 3/4 & 0 & 2/4 & A046 & 1 & 2/2 & 1 & 2/2 \\
A008 & 0 & 0/4 & 1 & 4/4 & A047 & 1 & 2/2 & 1 & 2/2 \\
A009 & 0 & 1/3 & 1 & 3/3 & A048 & 1 & 3/3 & 1 & 3/3 \\
A010 & 0 & 0/5 & 1 & 5/5 & A049 & 0 & 0/4 & 1 & 4/4 \\
A011 & 1 & 7/7 & 0 & 4/7 & A050 & 0 & 0/3 & 1 & 3/3 \\
A012 & 0 & 0/5 & 0 & 1/5 & A051 & 1 & 5/5 & 1 & 5/5 \\
A013 & 0 & 2/4 & 1 & 4/4 & A052 & 0 & 0/3 & 1 & 3/3 \\
A014 & 0 & 1/3 & 0 & 1/3 & A053 & 1 & 7/7 & 1 & 7/7 \\
A015 & 0 & 1/4 & 0 & 1/4 & A054 & 1 & 1/1 & 1 & 1/1 \\
A016 & 0 & 1/4 & 1 & 4/4 & A055 & 0 & 1/3 & 0 & 1/3 \\
A017 & 0 & 2/3 & 1 & 3/3 & A056 & 0 & 0/4 & 0 & 0/4 \\
A018 & 0 & 0/4 & 0 & 1/4 & A057 & 0 & 0/3 & 1 & 3/3 \\
A019 & 0 & 4/6 & 0 & 3/6 & A058 & 1 & 3/3 & 1 & 3/3 \\
A020 & 0 & 0/2 & 1 & 2/2 & A059 & 0 & 1/3 & 0 & 0/3 \\
A021 & 0 & 1/3 & 1 & 3/3 & A060 & 0 & 1/3 & 1 & 3/3 \\
A022 & 0 & 1/5 & 1 & 5/5 & A061 & 0 & 0/3 & 0 & 0/3 \\
A023 & 1 & 6/6 & 1 & 6/6 & A062 & 0 & 0/2 & 1 & 2/2 \\
A024 & 0 & 0/2 & 1 & 2/2 & A063 & 1 & 2/2 & 1 & 2/2 \\
A025 & 0 & 0/7 & 0 & 0/7 & A064 & 0 & 0/1 & 1 & 1/1 \\
A026 & 1 & 5/5 & 0 & 1/5 & A065 & 1 & 2/2 & 1 & 2/2 \\
A027 & 0 & 0/4 & 0 & 0/4 & A066 & 0 & 0/2 & 0 & 0/2 \\
A028 & 1 & 7/7 & 0 & 1/7 & A067 & 1 & 1/1 & 0 & 0/1 \\
A029 & 0 & 0/4 & 0 & 0/4 & A068 & 0 & 0/3 & 0 & 0/3 \\
A030 & 0 & 0/4 & 0 & 3/4 & A069 & 1 & 2/2 & 1 & 2/2 \\
A031 & 0 & 0/4 & 0$^{\dagger}$ & 0/4 & A070 & 1 & 1/1 & 1 & 1/1 \\
A032 & 0 & 0/4 & 0 & 0/4 & A071 & 0 & 0/1 & 1 & 1/1 \\
A033 & 0 & 2/3 & 0 & 2/3 & A072 & 0 & 1/3 & 0 & 0/3 \\
A034 & 0 & 1/5 & 0 & 0/5 & A073 & 0 & 1/3 & 0 & 0/3 \\
A035 & 0 & 0/6 & 0 & 1/6 & A074 & 1 & 1/1 & 1 & 1/1 \\
A036 & 0 & 0/3 & 0 & 0/3 & A075 & 1 & 3/3 & 0 & 0/3 \\
A037 & 0 & 3/4 & 0 & 1/4 & A076 & 0 & 0/2 & 0 & 0/2 \\
A038 & 0 & 1/2 & 1 & 2/2 & A077 & 0 & 0/1 & 0 & 0/1 \\
A039 & 1 & 4/4 & 1 & 4/4 & A078 & 0 & 2/3 & 0$^{\dagger}$ & 0/3 \\
\end{longtable}

\begin{longtable}{@{}lcccc@{\hspace{2.4em}}lcccc@{}}
\caption{Official per-case scores: partition B (131 cases)}\label{tab:case-scores-B}\\
\toprule
\textbf{Case} & \multicolumn{2}{c}{\textbf{D}} & \multicolumn{2}{c}{\textbf{U}} & \textbf{Case} & \multicolumn{2}{c}{\textbf{D}} & \multicolumn{2}{c}{\textbf{U}} \\
 & $s$ & $m/n$ & $s$ & $m/n$ & & $s$ & $m/n$ & $s$ & $m/n$ \\
\midrule
\endfirsthead
\multicolumn{10}{@{}l}{\small\itshape Partition B (continued)}\\
\toprule
\textbf{Case} & \multicolumn{2}{c}{\textbf{D}} & \multicolumn{2}{c}{\textbf{U}} & \textbf{Case} & \multicolumn{2}{c}{\textbf{D}} & \multicolumn{2}{c}{\textbf{U}} \\
 & $s$ & $m/n$ & $s$ & $m/n$ & & $s$ & $m/n$ & $s$ & $m/n$ \\
\midrule
\endhead
\midrule\multicolumn{10}{r@{}}{\small\itshape Continued on next page}\\
\endfoot
\bottomrule
\endlastfoot
B001 & 0 & 0/7 & 0 & 2/7 & B065 & 0 & 2/5 & 0 & 2/5 \\
B002 & 1 & 2/2 & 0 & 0/2 & B066 & 0 & 1/3 & 0 & 0/3 \\
B003 & 0 & 1/3 & 0 & 1/3 & B067 & 0 & 0/2 & 1 & 2/2 \\
B004 & 0 & 3/4 & 1 & 4/4 & B068 & 0 & 1/3 & 1 & 3/3 \\
B005 & 0 & 0/2 & 1 & 2/2 & B069 & 1 & 3/3 & 0 & 1/3 \\
B006 & 1 & 3/3 & 1 & 3/3 & B070 & 0 & 0/2 & 0 & 0/2 \\
B007 & 0 & 0/4 & 0 & 2/4 & B071 & 0 & 0/4 & 0 & 0/4 \\
B008 & 0 & 0/7 & 0 & 4/7 & B072 & 0 & 0/4 & 0 & 0/4 \\
B009 & 0 & 0/3 & 0 & 2/3 & B073 & 0 & 1/4 & 0 & 0/4 \\
B010 & 1 & 1/1 & 1 & 1/1 & B074 & 0 & 0/4 & 0 & 2/4 \\
B011 & 0 & 0/3 & 0 & 0/3 & B075 & 0 & 0/4 & 0 & 0/4 \\
B012 & 0 & 1/2 & 0 & 0/2 & B076 & 0 & 0/5 & 0 & 2/5 \\
B013 & 0 & 2/3 & 0 & 1/3 & B077 & 0 & 0/7 & 0 & 3/7 \\
B014 & 0 & 0/2 & 1 & 2/2 & B078 & 0 & 1/6 & 1 & 6/6 \\
B015 & 0 & 0/2 & 1 & 2/2 & B079 & 0 & 3/4 & 0 & 2/4 \\
B016 & 1 & 1/1 & 1 & 1/1 & B080 & 0 & 0/5 & 0 & 0/5 \\
B017 & 0 & 1/3 & 0 & 1/3 & B081 & 0 & 0/4 & 0 & 0/4 \\
B018 & 0 & 2/4 & 0 & 2/4 & B082 & 1 & 6/6 & 0 & 2/6 \\
B019 & 0 & 2/5 & 1 & 5/5 & B083 & 0 & 2/3 & 1 & 3/3 \\
B020 & 0 & 0/3 & 0 & 0/3 & B084 & 0 & 0/9 & 0 & 3/9 \\
B021 & 0 & 0/2 & 0 & 0/2 & B085 & 0 & 0/5 & 0 & 1/5 \\
B022 & 1 & 2/2 & 0 & 1/2 & B086 & 0 & 0/2 & 0 & 0/2 \\
B023 & 1 & 2/2 & 1 & 2/2 & B087 & 0 & 0/3 & 1 & 3/3 \\
B024 & 0 & 0/3 & 1 & 3/3 & B088 & 0 & 0/4 & 1 & 4/4 \\
B025 & 1 & 1/1 & 1 & 1/1 & B089 & 0 & 0/2 & 0 & 0/2 \\
B026 & 1 & 2/2 & 0 & 0/2 & B090 & 0 & 0/2 & 0 & 1/2 \\
B027\_a & 1 & 2/2 & 0 & 0/2 & B091 & 0 & 0/3 & 0 & 2/3 \\
B027\_b & 0 & 0/2 & 0 & 0/2 & B092 & 0 & 1/4 & 0 & 1/4 \\
B028 & 0 & 1/4 & 1 & 4/4 & B093 & 1 & 2/2 & 0 & 1/2 \\
B029\_a & 1 & 2/2 & 1 & 2/2 & B094 & 0 & 0/2 & 0 & 0/2 \\
B029\_b & 1 & 2/2 & 1 & 2/2 & B095 & 0 & 2/4 & 0 & 2/4 \\
B030 & 1 & 1/1 & 1 & 1/1 & B096 & 1 & 3/3 & 1 & 3/3 \\
B031 & 0 & 2/5 & 0 & 2/5 & B097 & 0 & 0/4 & 0 & 0/4 \\
B032 & 0 & 0/3 & 0 & 0/3 & B098 & 0 & 0/3 & 0 & 0/3 \\
B033 & 0 & 1/5 & 1 & 5/5 & B099 & 1 & 3/3 & 1 & 3/3 \\
B034 & 1 & 2/2 & 1 & 2/2 & B100 & 0 & 0/2 & 1 & 2/2 \\
B035 & 1 & 2/2 & 1 & 2/2 & B101 & 0 & 0/4 & 0 & 0/4 \\
B036 & 1 & 3/3 & 0 & 2/3 & B102 & 0 & 1/4 & 0 & 1/4 \\
B037 & 0 & 2/3 & 0 & 2/3 & B103 & 0 & 0/6 & 0 & 0/6 \\
B038 & 0 & 0/3 & 0 & 0/3 & B104 & 1 & 3/3 & 1 & 3/3 \\
B039 & 0 & 1/4 & 0 & 1/4 & B105 & 0 & 0/5 & 0 & 0/5 \\
B040 & 0 & 0/1 & 0 & 0/1 & B106 & 0 & 0/2 & 0 & 0/2 \\
B041 & 0 & 1/3 & 0 & 0/3 & B107 & 0 & 0/4 & 0$^{\dagger}$ & 0/4 \\
B042 & 0 & 0/3 & 0$^{\dagger}$ & 0/3 & B108 & 0 & 0/3 & 0 & 0/3 \\
B043 & 0 & 0/4 & 0$^{\dagger}$ & 0/4 & B109 & 1 & 3/3 & 0$^{\dagger}$ & 0/3 \\
B044 & 0 & 1/3 & 0 & 1/3 & B110 & 0 & 1/5 & 0 & 1/5 \\
B045 & 0 & 2/3 & 0 & 2/3 & B111\_a & 1 & 2/2 & 1 & 2/2 \\
B046 & 0 & 0/6 & 0 & 0/6 & B111\_b & 0 & 2/3 & 0 & 0/3 \\
B047 & 0 & 0/6 & 0 & 2/6 & B112 & 1 & 3/3 & 0 & 1/3 \\
B048 & 0 & 0/4 & 1 & 4/4 & B113 & 0 & 0/3 & 1 & 3/3 \\
B049 & 0 & 1/6 & 0 & 1/6 & B114 & 0 & 1/2 & 0 & 1/2 \\
B050 & 1 & 5/5 & 0 & 0/5 & B115 & 0 & 1/6 & 0 & 1/6 \\
B051 & 0 & 0/5 & 0 & 0/5 & B116 & 1 & 7/7 & 0 & 6/7 \\
B052 & 0 & 1/4 & 0 & 1/4 & B117 & 0 & 3/4 & 0 & 3/4 \\
B053 & 0 & 0/2 & 0 & 0/2 & B118 & 1 & 3/3 & 0 & 1/3 \\
B054 & 1 & 6/6 & 0 & 0/6 & B119 & 0 & 1/4 & 0 & 1/4 \\
B055 & 0 & 0/2 & 0 & 0/2 & B120 & 0 & 0/3 & 1 & 3/3 \\
B056 & 0 & 0/4 & 0 & 0/4 & B121 & 1 & 2/2 & 0 & 1/2 \\
B057 & 0 & 0/4 & 0 & 0/4 & B122 & 0 & 1/4 & 0 & 1/4 \\
B058 & 0 & 1/5 & 0 & 1/5 & B123 & 0 & 0/3 & 1 & 3/3 \\
B059 & 0 & 0/2 & 1 & 2/2 & B124 & 0 & 1/3 & 0 & 0/3 \\
B060 & 0 & 0/2 & 0 & 1/2 & B125 & 1 & 3/3 & 1 & 3/3 \\
B061 & 0 & 0/2 & 0 & 1/2 & B126 & 0 & 1/2 & 0 & 0/2 \\
B062 & 0 & 0/4 & 0 & 0/4 & B127 & 0 & 1/5 & 0$^{\dagger}$ & 0/5 \\
B063 & 0 & 0/2 & 0 & 0/2 & B128 & 0 & 0/6 & 0 & 0/6 \\
B064 & 0 & 1/4 & 0 & 1/4 &  &  &  &  &  \\
\end{longtable}

\begin{longtable}{@{}lcccc@{\hspace{2.4em}}lcccc@{}}
\caption{Official per-case scores: partition C (110 cases)}\label{tab:case-scores-C}\\
\toprule
\textbf{Case} & \multicolumn{2}{c}{\textbf{D}} & \multicolumn{2}{c}{\textbf{U}} & \textbf{Case} & \multicolumn{2}{c}{\textbf{D}} & \multicolumn{2}{c}{\textbf{U}} \\
 & $s$ & $m/n$ & $s$ & $m/n$ & & $s$ & $m/n$ & $s$ & $m/n$ \\
\midrule
\endfirsthead
\multicolumn{10}{@{}l}{\small\itshape Partition C (continued)}\\
\toprule
\textbf{Case} & \multicolumn{2}{c}{\textbf{D}} & \multicolumn{2}{c}{\textbf{U}} & \textbf{Case} & \multicolumn{2}{c}{\textbf{D}} & \multicolumn{2}{c}{\textbf{U}} \\
 & $s$ & $m/n$ & $s$ & $m/n$ & & $s$ & $m/n$ & $s$ & $m/n$ \\
\midrule
\endhead
\midrule\multicolumn{10}{r@{}}{\small\itshape Continued on next page}\\
\endfoot
\bottomrule
\endlastfoot
C001 & 0 & 2/3 & 0 & 2/3 & C055 & 0 & 1/2 & 0 & 0/2 \\
C002 & 0 & 1/3 & 0 & 2/3 & C056 & 0 & 0/5 & 0 & 0/5 \\
C003 & 0 & 0/2 & 0 & 0/2 & C057 & 0 & 1/2 & 0 & 0/2 \\
C004 & 0 & 0/4 & 0 & 0/4 & C058 & 0 & 0/2 & 0 & 0/2 \\
C005 & 0 & 3/7 & 0 & 2/7 & C059 & 0 & 0/3 & 0 & 1/3 \\
C006 & 0 & 1/6 & 0 & 1/6 & C060 & 0 & 0/3 & 0 & 0/3 \\
C007 & 0 & 0/5 & 1 & 5/5 & C061 & 0 & 1/5 & 0 & 0/5 \\
C008 & 0 & 2/3 & 0 & 2/3 & C062 & 0 & 3/5 & 0 & 3/5 \\
C009 & 0 & 0/3 & 0 & 2/3 & C063 & 0 & 1/6 & 0 & 1/6 \\
C010 & 0 & 0/5 & 0 & 0/5 & C064 & 0 & 2/3 & 0 & 0/3 \\
C011 & 1 & 2/2 & 1 & 2/2 & C065 & 1 & 4/4 & 0 & 2/4 \\
C012 & 0 & 1/3 & 0 & 2/3 & C066 & 0 & 1/6 & 0 & 0/6 \\
C013 & 0 & 0/4 & 0 & 0/4 & C067 & 0 & 0/3 & 0 & 0/3 \\
C014 & 0 & 2/7 & 0 & 4/7 & C068 & 0 & 0/6 & 0 & 0/6 \\
C015 & 0 & 0/4 & 0 & 0/4 & C069 & 0 & 0/6 & 0$^{\dagger}$ & 0/6 \\
C016 & 0 & 0/5 & 0 & 0/5 & C070 & 0 & 0/6 & 0 & 0/6 \\
C017 & 0 & 0/5 & 0 & 0/5 & C071 & 0 & 1/4 & 0 & 1/4 \\
C018 & 0 & 0/2 & 1 & 2/2 & C072 & 1 & 5/5 & 0 & 4/5 \\
C019 & 0 & 2/5 & 0 & 2/5 & C073 & 0 & 1/4 & 0 & 1/4 \\
C020 & 0 & 0/6 & 0$^{\ddagger}$ & 0/6 & C074 & 0 & 1/4 & 0 & 1/4 \\
C021 & 0 & 0/6 & 0 & 0/6 & C075 & 0 & 0/3 & 0 & 0/3 \\
C022 & 0 & 2/4 & 1 & 4/4 & C076 & 0 & 1/4 & 0$^{\dagger}$ & 0/4 \\
C023 & 0 & 1/5 & 0 & 1/5 & C077 & 0 & 1/3 & 0 & 1/3 \\
C024 & 0 & 2/4 & 0 & 2/4 & C078 & 0 & 3/4 & 0 & 2/4 \\
C025 & 0 & 0/4 & 0 & 0/4 & C079 & 0 & 0/5 & 0 & 0/5 \\
C026 & 0 & 0/5 & 0 & 0/5 & C080 & 0 & 2/6 & 0 & 2/6 \\
C027 & 0 & 0/8 & 0 & 1/8 & C081 & 0 & 0/1 & 0 & 0/1 \\
C028 & 0 & 0/5 & 0 & 4/5 & C082 & 0 & 1/2 & 0 & 1/2 \\
C029 & 0 & 1/6 & 0 & 1/6 & C083 & 1 & 4/4 & 0 & 0/4 \\
C030 & 0 & 2/7 & 0 & 1/7 & C084 & 0 & 0/5 & 0 & 4/5 \\
C031 & 0 & 1/4 & 0 & 1/4 & C085 & 0 & 1/6 & 0 & 2/6 \\
C032\_a & 1 & 1/1 & 1 & 1/1 & C086 & 0 & 0/4 & 0 & 2/4 \\
C032\_b & 0 & 1/2 & 1 & 2/2 & C087 & 0 & 0/6 & 0 & 0/6 \\
C033 & 0 & 1/4 & 1 & 4/4 & C088 & 0 & 0/7 & 0 & 2/7 \\
C034 & 0 & 2/4 & 0$^{\dagger}$ & 0/4 & C089 & 1 & 6/6 & 0 & 1/6 \\
C035 & 0 & 0/3 & 1 & 3/3 & C090 & 0 & 2/3 & 0 & 2/3 \\
C036 & 0 & 1/4 & 0 & 2/4 & C091 & 0 & 0/5 & 0 & 0/5 \\
C037 & 0 & 0/6 & 0 & 0/6 & C092 & 0 & 0/3 & 0 & 0/3 \\
C038 & 0 & 0/5 & 0 & 4/5 & C093 & 0 & 1/5 & 0 & 1/5 \\
C039 & 0 & 2/4 & 0 & 1/4 & C094 & 1 & 4/4 & 0 & 0/4 \\
C040 & 0 & 0/3 & 0 & 0/3 & C095 & 0 & 1/4 & 0 & 1/4 \\
C041 & 0 & 0/1 & 0 & 0/1 & C096 & 0 & 0/6 & 0 & 0/6 \\
C042 & 0 & 0/5 & 0 & 2/5 & C097 & 0 & 1/8 & 0 & 1/8 \\
C043 & 0 & 0/5 & 0$^{\dagger}$ & 0/5 & C098 & 0 & 1/2 & 1 & 2/2 \\
C044 & 0 & 2/8 & 0 & 2/8 & C099 & 0 & 0/8 & 0 & 2/8 \\
C045 & 0 & 1/5 & 0 & 2/5 & C100 & 0 & 1/5 & 0 & 1/5 \\
C046 & 0 & 0/3 & 0 & 0/3 & C101 & 0 & 0/3 & 0 & 0/3 \\
C047 & 0 & 5/7 & 0 & 1/7 & C102 & 0 & 0/3 & 0 & 2/3 \\
C048 & 0 & 0/4 & 0 & 0/4 & C103 & 0 & 1/5 & 0 & 1/5 \\
C049 & 0 & 4/5 & 1 & 5/5 & C104 & 1 & 3/3 & 0$^{\ddagger}$ & 0/3 \\
C050 & 0 & 1/5 & 0 & 2/5 & C105 & 0 & 3/4 & 1 & 4/4 \\
C051 & 0 & 4/10 & 0 & 1/10 & C106 & 0 & 1/5 & 0 & 1/5 \\
C052 & 1 & 7/7 & 0 & 0/7 & C107 & 0 & 0/2 & 0 & 0/2 \\
C053 & 0 & 0/2 & 0 & 0/2 & C108 & 0 & 4/5 & 0 & 1/5 \\
C054 & 0 & 2/10 & 0 & 2/10 & C109 & 0 & 3/5 & 0 & 1/5 \\
\end{longtable}

\endgroup

\end{document}